# Class Activation Mapping in Explainable Computer Vision: A Method-Centered Review of CNN, Transformer, and Foundation-Model-Era Visual Explanations

AmirHossein Eshghi[1]; Hamid Saadatfar[2]; Seyyed Ali Hoseini[3]; AmirMohsen Eshghi[4]; Siavash Arjomand Bigdeli[5*]

[1] M.Sc. student, Department of Computer Engineering, University of Birjand, Iran

Amir.eshghi@birjand.ac.ir

[2] Associate Professor, Department of Computer Engineering, University of Birjand, Iran

saadatfar@birjand.ac.ir

[3] Assistant professor, Department of Computer Engineering, University of Birjand, Iran

Sa.hoseini@birjand.ac.ir

[4] B.Sc. student, Department of Computer Engineering, University of Birjand, Iran

Amirmohseneshghi@birjand.ac.ir

[5] Associate Professor, Department of Applied Mathematics and Computer Science Visual Computing, Technical University of Denmark, Denmark

sarbi@dtu.dk (Corresponding Author)

## Abstract

Class activation mapping (CAM) is one of the most widely used visual explanation families in explainable artificial intelligence. Its purpose is intuitive: it converts internal model evidence into a heatmap that highlights the image regions, convolutional channels, tokens, or patches that support a target class or concept. Since the first CAM formulation in 2016, the field has moved far beyond global-average-pooled CNN classifiers. CAM-style methods now include gradient-based post-hoc explanations, gradient-free score and ablation methods, high-resolution upscaling, weakly supervised localization and segmentation, transformer token attribution, causal and debiasing methods, and foundation-model-era approaches that use CLIP, DINO, SAM, or feature-distribution comparisons. This review synthesizes a strict corpus of 57 method-centered papers published from 2016 onward. The paper develops a taxonomy that separates methods by attribution mechanism, architectural dependence, and evaluation objective. It then reviews gradient-based CAMs, recent and hybrid CAM-style methods, and model-based or architecture-aware methods. Across the corpus, the main trend is clear: the field is shifting from explaining one class score in one low-resolution CNN layer toward comparative, multi-layer, probabilistic, token-aware, and foundation-model-aware explanations. At the same time, evaluation remains fragmented. Faithfulness, localization, robustness, computational cost, and human trust are often measured with different protocols. The review therefore emphasizes not only what each method contributes, but also which gap it leaves open and which later methods attempt to close that gap.

**Keywords:** Class activation map, explainable AI, visual explanation, CNN, Vision Transformer

## 1. Introduction

Deep visual models have become central to image classification, segmentation, medical imaging, autonomous perception, industrial inspection, and many other decision systems. Their success comes from the ability to learn layered visual representations, but the same layered representation makes their decisions hard to interpret. In high-stakes settings, a correct class label is not enough. A user also needs to know whether the model used meaningful evidence, whether it relied on a background shortcut, and whether its reasoning is stable under changes in the image. CAM-style methods address this problem by producing visual evidence maps. A heatmap is not a complete explanation of a neural network, but it gives a compact and human-readable view of where the model looked when it produced a target output.

The original CAM paper showed that a classification-trained CNN can be made localizable when global average pooling connects spatial feature maps to class scores [1]. Grad-CAM then made the idea more general by using target-class gradients, so that CAM-like maps could be generated for ordinary CNN classifiers and even CNN-based multimodal models [2]. This shift changed CAM from a special architectural property into a practical post-hoc explanation tool. The field then expanded in several directions. Some methods refined gradient weighting to improve object coverage [3], while others avoided gradients and measured channel importance through forward score changes or ablations [8], [9]. A separate stream used CAMs as seeds for weakly supervised localization and segmentation, where the goal is not only to explain a model but also to train a downstream dense predictor [4]-[7], [11]. More recent work extends CAM-style attribution to transformers, class tokens, patch tokens, visual prompts, CLIP prompts, DINO affinity graphs, and SAM-generated masks [30]-[33], [41]-[43], [47]-[50].

This review is method-centered. It does not include papers that merely use Grad-CAM as a visualization figure in an application study. Instead, a paper must generate, refine, evaluate, or theoretically analyze a CAM-style map. This scope is important because CAM is now used in many application papers, but application-only usage does not necessarily change the method. The contribution of this review is to organize the method literature into a coherent development story: from CNN-specific localization, to gradient and score-based attribution, to high-resolution and causal refinements, to transformer and foundation-model-era explanations.

The review has four goals. First, it defines a reproducible search and inclusion protocol for a strict corpus of 57 papers. Second, it clarifies the main technical terms and evaluation metrics used in CAM research. Third, it develops a taxonomy that separates gradient-based, gradient-free, hybrid, high-resolution, token-level, foundation-model-era, and architecture-aware methods. Fourth, it compares representative results where the corpus reports compatible metrics on common datasets. Because CAM evaluation is not standardized, the quantitative figures in this review are deliberately limited to values that are directly reported in the included papers and are interpreted with their protocol constraints.

## 1.1 Search and inclusion protocol

The literature search followed a method-centered protocol. The primary databases and libraries were IEEE Xplore and IEEE/CVF Open Access, ACM Digital Library, Elsevier ScienceDirect, SpringerLink, and Springer Nature; venue-level cross-checking was performed only within these selected sources. The search was restricted to 2016 onward because the original CAM formulation was published in 2016 [1]. The strict venue and publisher filter focused on IEEE/CVF, ACM, Elsevier, Springer/Springer Nature, CVPR, ICCV, ECCV, WACV, ICASSP, ICIP, ACM MM, CHI, ACCV, and related journals represented in the corpus.

**Table 1. Search strings, inclusion rules, and exclusion rules used to build the strict CAM-style corpus.**

| Block | Boolean syntax used in screening | Purpose |
|---|---|---|
| Concept block | ("class activation map" OR CAM OR "Grad-CAM" OR "Score-CAM" OR "Ablation-CAM" OR "LayerCAM" OR "activation map" OR "token attention map") | Find CAM-style heatmap and activation-map papers. |
| Architecture block | (CNN OR "convolutional neural network" OR "vision transformer" OR ViT OR CLIP OR "Segment Anything Model" OR "foundation model") | Restrict the search to visual architectures relevant to computer vision XAI. |
| Method block | (explainability OR interpretability OR attribution OR saliency OR "weakly supervised localization" OR "weakly supervised semantic segmentation") | Remove papers unrelated to explanation, attribution, WSOL, or WSSS. |
| Venue/year filter | 2016 onward AND (Elsevier OR Springer OR ACM OR IEEE OR CVPR OR ICCV OR ECCV OR WACV OR ICASSP OR ICIP OR ACCV) | Keep the corpus method-centered and venue/publisher-bounded. |
| Inclusion rule | Method must generate, refine, evaluate, or theoretically analyze CAM-style heatmaps, activation maps, token maps, or patch attribution maps. | Defines the strict corpus. |

| Block | Boolean syntax used in screening | Purpose |
|---|---|---|
| Exclusion rule | Application-only Grad-CAM use, survey-only papers, non-vision papers, pre-2016 papers, duplicates, and records outside the selected venue/publisher scope were excluded. | Prevents the review from becoming an application survey. |

Table 1 is central to the scope of the review. It explains why a paper such as Score-CAM is included, because it changes how CAM weights are computed [8], while an application paper that only shows a Grad-CAM visualization is not included. Figure 1 gives the same process as a formal flow diagram. The diagram is PRISMA-inspired, but it is adapted to a method-centered computer-vision review: the final exact count is the strict corpus of 57 papers, while the intermediate steps are described operationally rather than as clinical-trial-style counts.

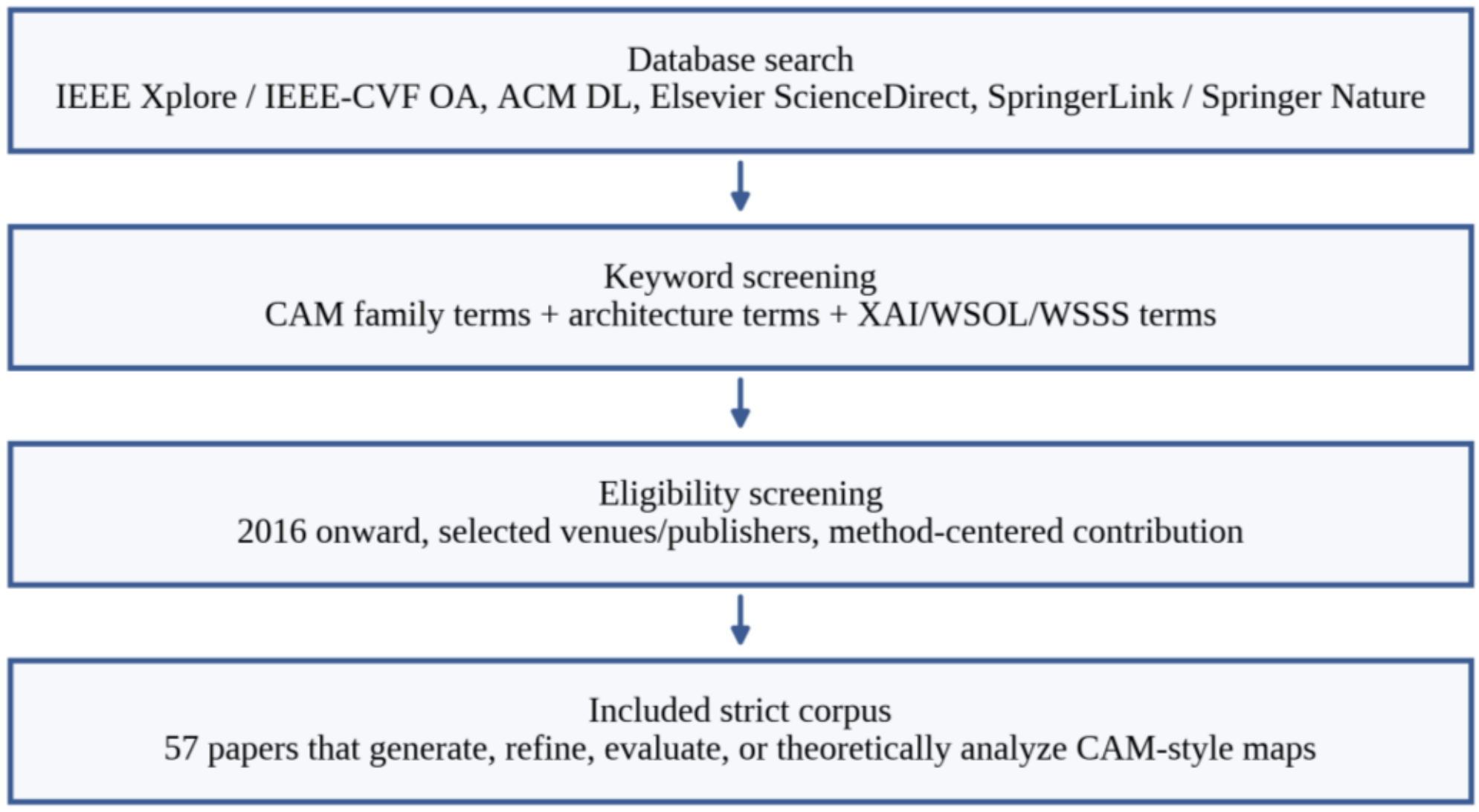

**Figure 1. Method-centered search and inclusion protocol. The final corpus includes 57 papers that generate, refine, evaluate, or theoretically analyze CAM-style explanations.**

The rest of the article is organized as follows. Section 2 introduces the background, notation, and evaluation metrics. Section 3 presents the taxonomy. Section 4 reviews gradient-based CAM methods for CNNs and transformers. Section 5 reviews recent and hybrid CAM-style methods, including gradient-free, high-resolution, causal, debiasing, and foundation-model-era work. Section 6 compares method families by cost, faithfulness, and localization trends. Section 7 discusses model-based and architecture-aware CAM methods. Sections 8-10 provide comparative discussion, open challenges, and conclusions.

## 2. Background and Definitions

A CAM-style explanation begins with a target output. In a closed-set classifier this target is usually a class logit or class probability. In a detector it may be a box or class head. In a transformer it may be a class token or text-conditioned similarity score. In a representation model it may be a feature vector rather than a class score. Let $A^k$ be the kth feature map at a selected layer and $y^c$ be the score for class c. The common CAM form is $L^c = h(\sum_k \alpha_k^c A^k)$, where $\alpha_k^c$ is the importance of the k-th map and h is usually a non-negative activation such as ReLU. Most CAM papers differ in how they estimate $\alpha_k^c$, where they extract $A^k$, and how they evaluate the resulting map. The original CAM method obtains $\alpha_k^c$ directly from the classifier weight connected to global-average-pooled feature map k [1]. Grad-CAM replaces this architectural weight with the average gradient of the target score with respect to the feature map [2]. Score-CAM avoids gradients by masking the input with upsampled activation maps and using the forward score as the weight [8]. Ablation-CAM measures the effect of removing feature maps [9]. LayerCAM uses positive gradients at individual spatial locations, making it possible to collect fine information from shallower layers [16]. Finer-CAM changes the target itself: instead of explaining only why class c is high, it explains why class c is higher than a similar reference class [49].

In transformers, the spatial feature map is replaced by patch tokens, class tokens, prompt tokens, or attention matrices. This change is not cosmetic. In CNNs, locality is built into convolution; in ViTs, long-range relations are carried by attention. Transformer CAM-style methods therefore need to decide whether explanation should follow attention weights, relevance propagation, class-to-patch attention, or learned prompt attention [30]-[33], [41], [48]. In foundation models the target can become even more flexible. CLIP explanations depend on the text prompt [47], DINO-based methods use self-supervised semantic affinity [43], and SAM-assisted methods inject segmentation priors into CAM training [42].

**Table 2. Core definitions used in the review.**

| Term | Definition |
|---|---|
| CAM-style map | A spatial, token-level, or patch-level map that explains a target class, concept, prompt, detector output, or feature representation. |
| Attribution weight | The coefficient $\alpha_k^c$ used to combine feature maps, channels, tokens, or regions. |
| Gradient-based CAM | Uses derivatives of a target score with respect to features or attention to compute explanation weights. |
| Gradient-free CAM | Uses forward scores, ablations, masks, PCA, clustering, perturbations, or ensembles without relying on gradients. |
| High-resolution CAM | Aims to recover fine spatial details through layer fusion, guided upscaling, augmentation, region integration, or comparison. |
| Token/patch CAM | Explains ViTs or hybrid models through class tokens, patch tokens, attention maps, or prompts. |
| Faithfulness | Measures whether highlighted evidence is truly important to the model output. |
| Localization | Measures whether highlighted evidence aligns with boxes, masks, object parts, or expert annotations. |
| WSOL/WSSS | Weakly supervised object localization and semantic segmentation, where CAMs are often used as pseudo-label seeds. |

Table 2 is useful because CAM research now covers more than CNN heatmaps. The same word 'activation' can refer to a convolutional channel, a transformer patch token, a detector head, a CLIP embedding, or a SAM-guided segment. Without these definitions, comparisons between methods become ambiguous.

## 2.1 Evaluation criteria

Evaluation is a persistent source of disagreement in CAM research. A visually sharp map can be unfaithful, and a faithful map can reveal a model shortcut that humans do not want. For this reason, the corpus uses multiple evaluation families. Perturbation metrics such as deletion and insertion test whether removing or restoring salient pixels changes the prediction. Average Drop and Increase in Confidence measure how confidence changes when only the highlighted region is retained. ROAD and ADCC combine perturbation quality with complexity or coherence. Localization metrics compare maps with boxes, masks, or expert regions. In WSSS, mIoU and DSC measure whether CAM-derived pseudo-labels are useful for segmentation. Human-centered studies evaluate whether explanations help users calibrate trust or complete a task [2], [44].

**Table 3. Evaluation metrics used across the CAM-style literature.**

| Metric | Interpretation |
|---|---|
| Deletion | Remove the most salient pixels first; a faster confidence fall or lower AUC indicates stronger evidence removal. |
| Insertion | Start from a blurred/blank image and insert salient pixels; faster confidence recovery or higher AUC is better. |
| Average Drop / Increase | Confidence change when the highlighted evidence is retained. Used widely in Score-CAM-style papers. |
| ADCC | A composite score balancing average drop, coherence, and complexity; used in recent CAM comparisons [40], [52]. |
| ROAD | Remove-and-debias perturbation score used for robust saliency and CAM ensembles [54]. |
| Top-1 / Top-5 localization | A prediction is correct only if class prediction and box overlap satisfy the benchmark rule. |

| Metric | Interpretation |
|---|---|
| mIoU / DSC | Dense mask quality for WSSS and medical WSSS; higher values indicate better pseudo-label or segmentation quality. |
| Pointing game / energy localization | Checks whether maximum or high-energy activation lies inside annotated target regions. |
| Sanity check | Tests whether explanations change when model weights or labels are randomized. |

Table 3 also explains why this review avoids mixing incompatible numbers. A WSSS mIoU score cannot be directly compared with a deletion AUC, and a CLIP prompt-localization BoxAcc cannot be directly compared with CUB Top-1 localization accuracy. The quantitative figures below therefore use shared or clearly labeled protocols.

## 3. Taxonomy of CAM-style Methods

The taxonomy groups papers by their primary methodological contribution. Figure 2 shows the eight sections and their paper counts. The largest sections are gradient-free/perturbation CAMs, CNN-based explanations, high-resolution CAMs, and attention-gradient hybrids. This distribution reflects the history of the field. Early work asked how to localize evidence in CNNs. Later work asked how to make the map more faithful, less dependent on gradients, more spatially detailed, more useful for dense prediction, and more suitable for transformer and foundation architectures.

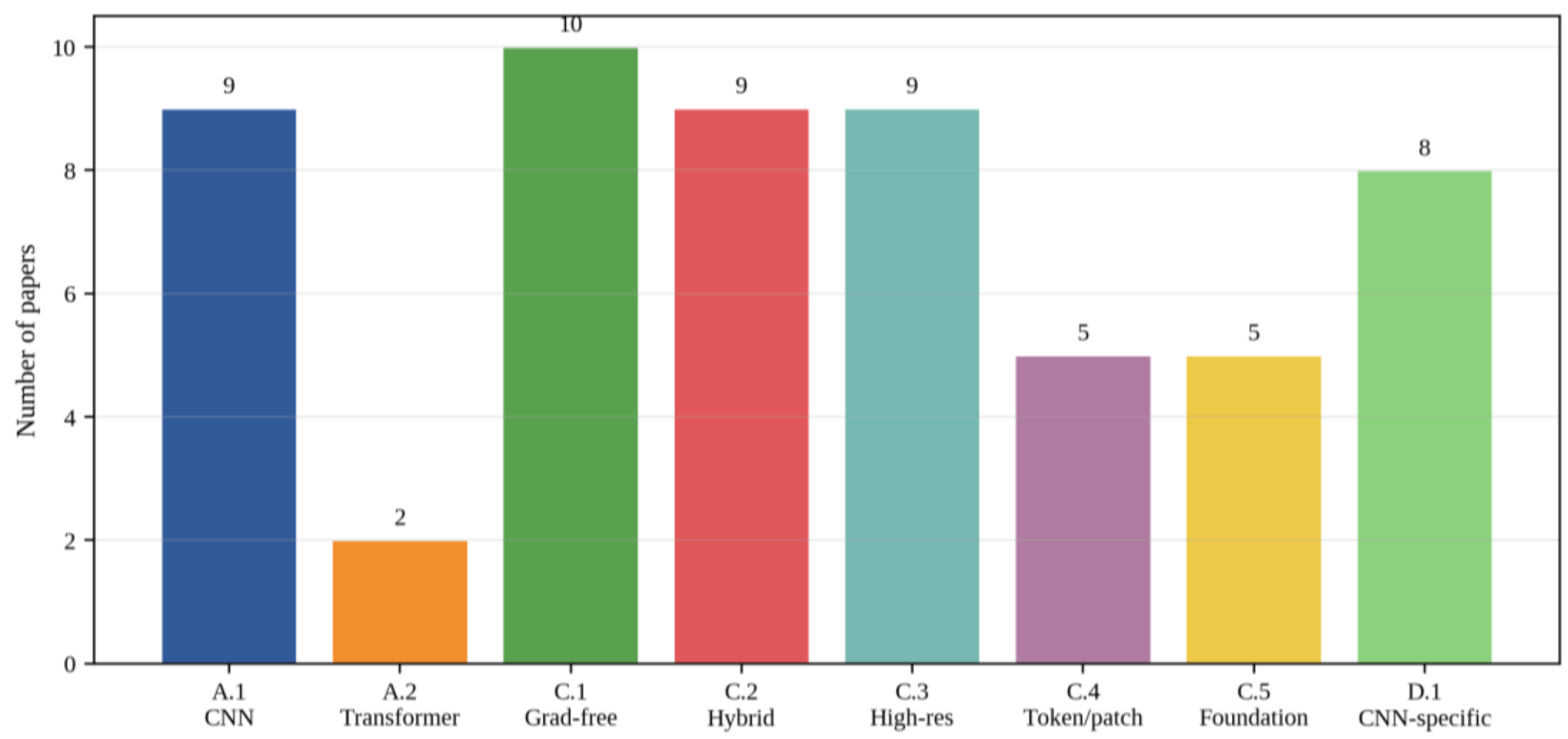

**Figure 2. Distribution of the strict 57-paper corpus across the eight primary taxonomy sections.**

Table 4 shows that the field cannot be ranked along one dimension. A gradient-based method may be fast and flexible but coarse. A gradient-free method may be more faithful but expensive. A high-resolution method may be better for pixel-level pseudo labels but less clearly tied to the model's causal mechanism. A foundation-model method may localize open-vocabulary concepts but depend strongly on prompt wording or pretrained priors. This is why the review first discusses method families separately and only then compares them.

**Table 4. Taxonomy of CAM-style methods and the main difference between categories.**

| Category | Main mechanism | Representative papers | Best use | Recurring gap |
|---|---|---|---|---|
| A.1 CNN-based explanations | Gradient, relevance, Shapley, or feature-vector explanations for CNNs. | Grad-CAM [2], Grad-CAM++ [3], Relevance-CAM [13], LIFT-CAM [15], FAM [28], ShapleyCAM [55], SIS-CAM | General post-hoc explanation and model debugging. | Gradient saturation, low spatial resolution, and uncertain faithfulness. |
| A.2 Transformer-based explanations | Relevance propagation through self-attention, co-attention, or encoder-decoder attention. | Transformer attribution [32], Generic Attention Explainability [33]. | ViT, VQA, image-text, and encoder-decoder models. | Attention alone is not explanation; residual and multimodal paths complicate attribution. |

| Category | Main mechanism | Representative papers | Best use | Recurring gap |
|---|---|---|---|---|
| C.1 Gradient-free / perturbation CAMs | Use forward scores, ablations, PCA, clustering, masking, probabilistic maps, or spatial feature perturbation. | Score-CAM [8], Ablation-CAM [9], Eigen-CAM [21], Cluster-CAM [38], CAPE [39], ReciproCAM [40], ScoreCAM++ [51]. | Avoid unstable gradients and support post-deployment settings. | Often more costly than gradient methods or sensitive to mask design. |
| C.2 Attention-gradient hybrid methods | Combine CAM with self-supervision, anti-adversarial manipulation, causality, contrast, or debiasing. | SEAM [11], AdvCAM [23], C2AM [25], C-CAM [26], PCAA [29], Debiased-CAM [44], CI-CAM [45]. | WSSS, WSOL, medical segmentation, and robust explanations. | Can be training-heavy and task-specific. |
| C.3 High-resolution CAMs | Improve spatial detail using augmentation, layer fusion, decoders, region integration, or contrastive comparison. | Augmented Grad-CAM [12], LayerCAM [16], F-CAM [18], Augmented Score-CAM [20], OLM [24], RPIM [46], Poly-CAM [37], Finer-CAM [49], GFR-CAM [52]. | Fine boundaries, pixel-level pseudo labels, and fine-grained traits. | High resolution can amplify noise if semantic targeting is weak. |
| C.4 Token-level / patch-level CAMs | Use class-to-patch attention, multiple class tokens, class-token infusion, or prompt attention. | TS-CAM [31], MCTformer [30], TransCAM [34], CTI [41], Prompt-CAM [48]. | ViT WSOL/WSSS and fine-grained analysis. | Class specificity, background control, and token-to-pixel alignment remain difficult. |
| C.5 Foundation-model-era explainability | Use CLIP, SAM, DINO, feature-distribution differences, or CAM ensembles. | S2C [42], ECA/DINO semantic guider [43], gScoreCAM [47], DiffCAM [50], MetaCAM [54]. | Open-vocabulary, self-supervised, segmentation-prior, and ensemble explanation. | Prompt sensitivity, pretrained bias, and mixed-source evidence. |
| D.1 CNN-specific methods | Modify CNN training, pooling, erasing, guidance, or reactivation to improve CAM-derived localization. | CAM [1], Hide-and-Seek [4], ACoL [5], SPG [6], ADL [7], Rethinking CAM [10], CREAM [27], LPCAM [35]. | WSOL and improved object completeness. | Often tied to classifier training assumptions and threshold choices. |

## 4. Gradient-based CAM Methods

Gradient-based methods are the reference point for most CAM research. They are attractive because they usually need one forward pass and one backward pass. They are also flexible: a target can be a class score, a caption score, a VQA answer, or another differentiable output. Their weakness is that gradients measure local sensitivity, not necessarily causal contribution. They can vanish, saturate, become noisy in shallow layers, or highlight features that change the score quickly but are not semantically meaningful. The development of gradient-based CAMs is therefore a sequence of attempts to preserve the speed of Grad-CAM while improving resolution, stability, theoretical grounding, or architectural reach.

### 4.1 CNN gradient, relevance, and game-theoretic CAMs

Grad-CAM [2] made CAM practical for ordinary CNNs. Instead of requiring a global-average-pooling architecture, it computes the importance of each feature map by averaging the gradient of the target score with respect to that map. The weighted feature maps are then summed and passed through ReLU. This design explains why Grad-CAM became widely used, it is simple, model-local, class-discriminative, and can be applied to CNN classifiers, captioning models, VQA systems, and other differentiable visual models. Its limitation is that the final convolutional layer is spatially coarse, and the average gradient can miss localized evidence or become weak when the model is saturated. Grad-CAM++ [3] addresses the first limitation by changing the gradient weighting rule. It uses a weighted combination of positive partial derivatives, so pixels that make strong positive contributions receive more emphasis. This improves multiple-object and multiple-instance explanation because the map is no longer controlled only by a single averaged gradient. The method is still derivative-based, however, so it does not remove the broader problem of noisy or saturated gradients. Integrated Grad-CAM [17] targets that problem from another direction. It computes a path integral of Grad-CAM-style scores, so the explanation is less dependent on the local gradient at only one input point. On PASCAL VOC 2007, the method reported strong EBPG and bounding-box localization values for ResNet-50, but the improvement comes with extra gradient evaluations and with the usual sensitivity of integrated methods to path design. SIS-CAM [56] addresses the noisy gradient problem by iteratively refining saliency maps. It squares gradients, fuses intermediate feature maps with the input image, and applies boundary-aware masking to emphasize critical regions. This iterative approach improves both faithfulness (deletion/insertion metrics) and spatial resolution, making SIS-CAM effective for high-resolution CAM generation in WSOL and WSSS tasks.

Relevance-CAM [13] is motivated by the observation that gradients become unreliable in intermediate layers. The paper uses layer-wise relevance propagation to derive the weighting components instead of ordinary gradients. This makes the explanation more stable in shallow and middle layers, where Grad-CAM may suffer from shattered gradients. The important difference is that Relevance-CAM is not trying only to sharpen the final-layer map; it tries to make class-specific maps reliable across layer depth. In the reported ResNet-50 evaluation, Relevance-CAM improved shallow-layer Average Drop and Average Increase relative to Grad-CAM and Grad-CAM++ [13]. Its remaining limitation is that relevance propagation depends on propagation rules, so the method is less plug-and-play than a standard backward gradient. LIFT-CAM [15] asks a more theoretical question: why should the coefficients of a CAM linear combination be chosen by a heuristic The method formulates each activation map as a feature in an additive attribution model and uses a DeepLIFT-based approximation to estimate SHAP-like coefficients. This gives the CAM weights a clearer additive-attribution meaning while keeping the method efficient. The paper reports higher insertion AUC and lower deletion AUC than Grad-CAM and Grad-CAM++ on ImageNet samples [15]. The main gap is that DeepLIFT-style attribution still depends on reference activations, and the SHAP connection is approximate rather than exact.

FAM [28] broadens the target of explanation. In tasks such as person re-identification or self-supervised representation learning, a test image may not correspond to a training class. Explaining a class logit is then less meaningful. FAM defines Feature Activation Mapping and explains which image regions contribute to the feature vector itself. This is an important shift because many modern vision systems use encoders as representation extractors rather than closed-set classifiers. The limitation is that the target feature must be carefully defined, and the resulting explanation is about representation formation rather than a single class decision. Abs-CAM [57], SSG-CAM [53], and ShapleyCAM [55] further illustrate how the gradient signal can be refined. Abs-CAM turns gradients into absolute positive contributions and fuses the resulting saliency with the input image, which reduces negative-gradient noise and improves deletion, insertion, and pointing-game behavior [57]. SSG-CAM uses smoothed second-order gradients and combines them with differential evolution to select and fuse the best multi-layer feature maps [53]. It directly addresses two weaknesses of first-order CAMs: saturation and manual layer selection. ShapleyCAM [55] builds a content-reserved game-theoretic framework and derives Shapley-style CAM weights by using gradients and Hessian information. It bridges scalable CAMs and fair feature attribution, but it also introduces second-order computation and utility-function design choices. Together, these methods show that the gradient family is moving from simple sensitivity toward more principled attribution.

## 4.2 Transformer gradient and relevance explanations

Transformer explanations require a different interpretation model from CNN explanations. In CNNs, spatial evidence is naturally organized in feature maps. In vision transformers, evidence flows through patch tokens, class tokens, residual connections, attention heads, and sometimes cross-modal attention. A naive attention map is therefore not automatically an explanation. Attention indicates how tokens exchange information, but it does not by itself show which tokens are relevant to the final decision. The two transformer papers in this group are important because they explicitly move beyond attention visualization and treat transformer explanation as a relevance-propagation problem.

Chefer, Gur, and Wolf [32] proposed a method, in that method starts from the observation that common transformer visualizations either display attention weights from one layer or roll attention matrices across layers. Both approaches can be misleading. A token can receive high attention without being important for the final class, and simple attention rollout assumes a linear flow of information that ignores nonlinear transformations, residual paths, and the difference between attention and relevance. The proposed method assigns local relevance using the Deep Taylor Decomposition principle and then propagates relevance through the transformer layers while accounting for attention and skip connections. The aim is to preserve the total amount of relevance as it moves backward through the model. In vision tasks, the resulting relevance scores can be mapped back to image patches and visualized as a class-specific explanation map. The method was benchmarked on visual transformer networks and a text classification task and was shown to outperform attention-only baselines. Its main limitation is that it requires access to model internals and careful bookkeeping through transformer components. It is therefore more principled than attention rollout but also more architecture-aware. The same authors then generalized the approach to bi-modal and encoder-decoder transformers, covering self-attention, co-attention, and encoder-decoder attention [33]. These papers are important because they show that transformer explanation is not simply CNN CAM with a different feature map. Relevance must move through token mixing, residual paths, and cross-modal interactions.

## 4.3 Quantitative observations for gradient-based CAMs

The gradient-based papers do not all use the same benchmark. Therefore, Figure 3 reports two within-protocol comparisons rather than mixing values into one artificial ranking. The left panel follows the Relevance-CAM ResNet-50 setting and shows how relevance weighting improves faithfulness, especially in Average Increase. The right panel follows the LIFT-CAM ImageNet setting and shows that LIFT-CAM improves insertion and deletion AUC relative to Grad-CAM and Grad-CAM++ [15].

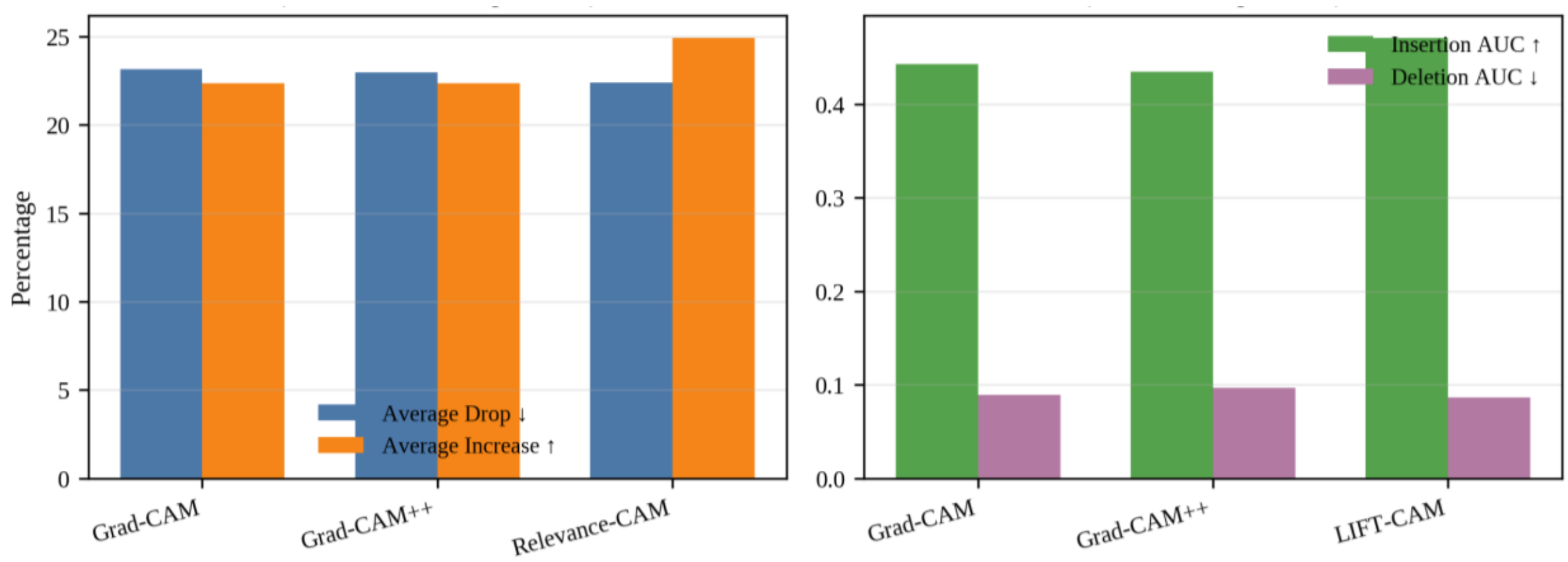

**Figure 3. Faithfulness observations for gradient-based and relevance-based CAMs. Each panel uses values reported under its own shared protocol; cross-panel values are not meant to be directly ranked.**

Figure 4 focuses on localization. Integrated Grad-CAM improves the ResNet-50 bounding-box score on PASCAL VOC 2007, while LIFT-CAM places a larger share of explanation energy inside ImageNet bounding boxes. The important point is not that one method is universally superior, but that each method improves a different part of the gradient-CAM pipeline: Integrated Grad-CAM improves sensitivity-aware scoring, while LIFT-CAM improves coefficient attribution.

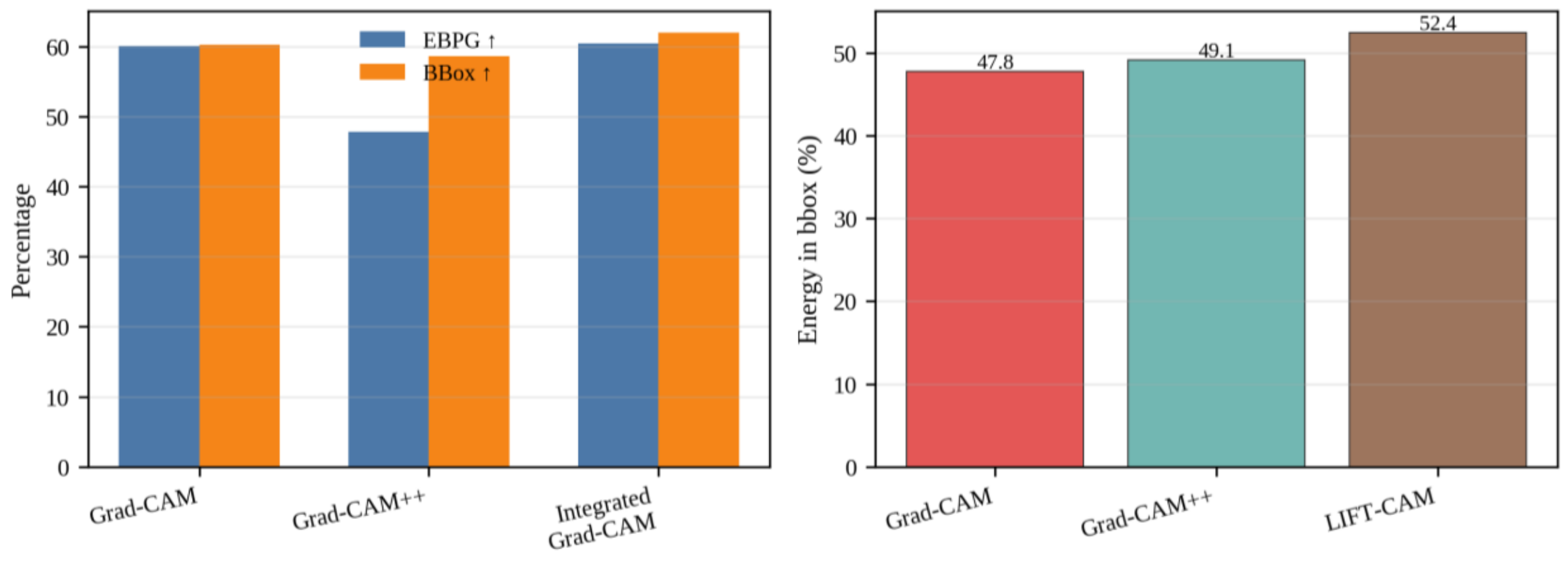

**Figure 4. Localization observations for gradient-based methods. The left panel uses PASCAL VOC 2007; the right panel uses ImageNet energy**

**Table 5. Summary of gradient-based, relevance-based, and game-theoretic CAM methods.**

| Method | Core mechanism | Innovation | Remaining gap | Datasets / results |
|---|---|---|---|---|
| Grad-CAM [2] | Average target gradients weight feature maps. | General post-hoc CAM for CNN and CNN-based multimodal models. | Coarse maps and gradient saturation. | ILSVRC, CUB, VQA, captioning; improved localization and trust calibration. |
| Grad-CAM++ [3] | Positive higher-order partial derivatives. | Better multiple-object and part coverage. | Still gradient- and final-layer-dependent. | ImageNet, PASCAL VOC, action recognition; improved over Grad-CAM in localization. |
| Integrated Grad-CAM [17] | Path-integrated gradient scoring. | Reduces sensitivity underestimation from saturation. | Baseline/path choices and extra computation. | Object localization and interpretation experiments. |
| Relevance-CAM [13] | Layer-wise relevance propagation weights. | Better intermediate-layer explanation and high-resolution maps. | Rule choices affect relevance propagation. | Localization and recognition evaluations across layers. |
| LIFT-CAM [15] | DeepLIFT approximation of SHAP-style activation-map weights. | Theoretical additive attribution basis for CAM coefficients. | Reference dependence and approximation error. | ImageNet, VOC, COCO, VQA; faster than exact SHAP-like computation. |
| FAM [28] | Explains feature vectors rather than class logits. | Extends CAM to representation models and Re-ID. | Requires defining a feature-level target. | Person Re-ID and self-supervised representation analysis. |
| Abs-CAM [57] | Absolute gradient optimization and image-map fusion. | Reduces noisy negative gradients. | Can blur evidence for vs. against a class. | Deletion, insertion, pointing game; improved localization clarity. |
| SIS-CAM [56] | Squared gradients plus iterative mask fusion. | Security-oriented saliency refinement. | More iterative cost than simple Grad-CAM. | ILSVRC2012val; deletion/insertion, average drop, sanity checks. |
| ShapleyCAM [55] | Game-theoretic CAM with gradients and Hessian approximation. | Links CAMs to Shapley-value fairness. | Second-order computation and utility design. | ImageNet across multiple networks; improved theoretical grounding. |
| SSG-CAM [53] | Second-order gradients, smoothing, and differential-evolution layer fusion. | Automatic multi-layer feature fusion. | Higher optimization cost. | WSOL, segmentation, biomedical lesion localization. |
| Transformer attribution [32] | Relevance propagation through attention and residuals. | Beyond attention visualization in ViTs. | Requires transformer-specific relevance bookkeeping. | ViT and text classification benchmarks. |
| Generic attention explainability [33] | Extends relevance to co-attention and encoder-decoder attention. | Explains multi-modal and generative transformer structures. | Complex cross-modal relevance paths. | VQA and encoder-decoder transformer tasks. |

Table 5 shows the internal logic of this family. Each new method keeps the basic CAM goal but changes the weighting signal: gradients, positive derivatives, path-integrated gradients, relevance, DeepLIFT, Shapley-style utility, or second-order gradients. The open gap is that no single weighting rule simultaneously guarantees causal faithfulness, high resolution, low cost, and model-agnostic deployment.

# 5. Recent and Hybrid CAM-style Methods

The second major family includes methods that either avoid gradients, combine CAMs with additional supervision, improve spatial resolution, or adapt CAMs to foundation models. This section follows the same pattern as Section 4: methods are described as a development story, each gap is stated, and later methods are positioned as partial responses.

## 5.1 Gradient-free and perturbation-based CNN methods

Score-CAM was a turning point because it removed the dependence on backpropagated gradients. It uses activation maps as masks, forwards masked images through the model, and uses the target-class confidence as the channel weight [8]. This often gives clean maps and passes sanity checks, but it is expensive because a separate forward pass may be needed for many channels. Ablation-CAM answers a related question by measuring how much the target score changes when a feature map is removed [9]. Its conceptual advantage is direct marginal contribution; its cost is repeated ablation. Ablation-CAM++ reduces this cost by grouping activation maps recursively, preserving the ablation idea while making it more time efficient [19].

Eigen-CAM uses principal components of learned feature representations instead of class-specific gradients or scores [21]. This makes it simple, class-independent, and robust when dense classifier layers fail, but it can lose explicit class discrimination. BBAM is different because it uses a trained object detector and finds the smallest image regions inside a bounding box that preserve detector behavior [22]. It is useful for weakly supervised semantic and instance segmentation with box labels, but it is detector-specific rather than a generic classifier explanation. AD-CAM, Cluster-CAM, CAPE, and ReciproCAM all respond to the same pressure: how can a gradient-free method be clearer without becoming too slow? AD-CAM uses lightweight spatial feature masks [36]. Cluster-CAM clusters feature maps to reduce the number of forward passes and combines cognition-base and cognition-scissors maps [38]. CAPE reformulates CAM as a probabilistic ensemble so that regional contributions can be meaningfully compared across classes [39]. ReciproCAM perturbs intermediate feature maps spatially and reports much faster runtime than Score-CAM while preserving strong ADCC performance [40].

ScoreCAM++ improves Score-CAM itself. It argues that min-max normalization can hide the difference between high- and low-priority activation values. By gating activation maps with a stronger activation function, especially tanh, ScoreCAM++ separates important and unimportant regions more clearly [51]. Figure 5 shows the ImageNet / VGG-19 comparison reported by ScoreCAM++. The method sharply lowers Average Drop and increases confidence relative to Score-CAM and gradient baselines. The result supports a broader lesson: even when the attribution family stays the same, normalization and gating can substantially affect faithfulness.

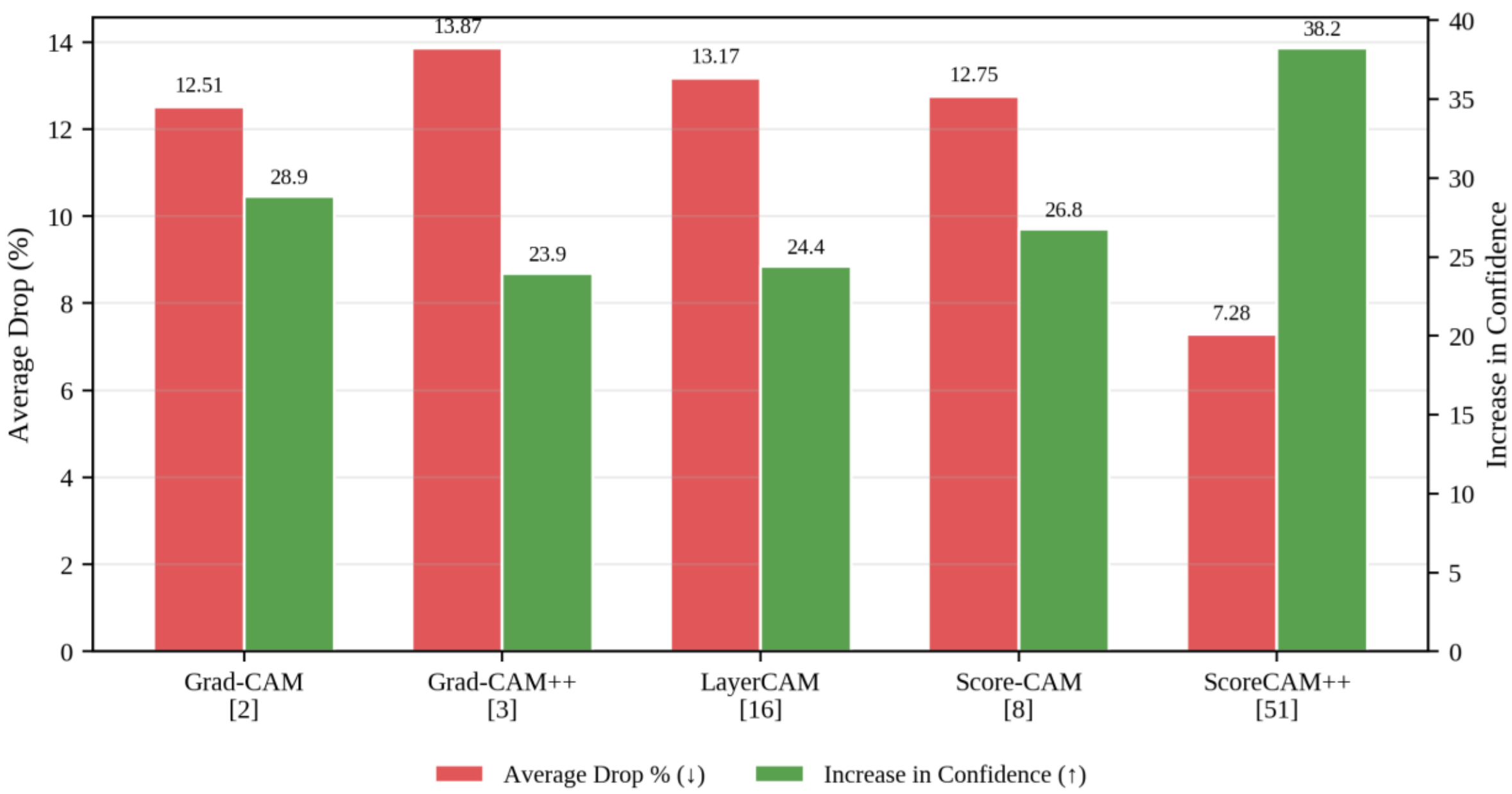

**Figure 5. ScoreCAM++ faithfulness results on ImageNet / VGG-19 using only methods from the strict corpus.**

## 5.2 Hybrid, causal, and weakly supervised methods

Hybrid CAM methods often use explanations as training signals. SEAM adds self-supervised equivariance: CAMs should transform consistently when the input is transformed [11]. Its pixel correlation module further refines pixels by similar neighbors, reducing under-activation and over-activation. AdvCAM takes a different route by anti-adversarially manipulating the input so that regions initially ignored by the classifier become involved in later attributions [23]. C2AM uses contrastive learning to generate class-agnostic activation maps from unlabeled data, separating foreground and background representations before using them to refine class-specific CAMs [25].

Causal and bias-aware methods address a deeper problem: a CAM can be faithful to a bad reason. CI-CAM uses causal intervention to reduce object-context entanglement in WSOL, especially when objects co-occur with backgrounds such as birds and branches or ducks and water [45]. C-CAM adapts this idea to medical WSSS, where organ co-occurrence and ambiguous boundaries make ordinary CAMs unreliable; it models category-causality and anatomy-causality chains and reports strong DSC improvements on ProMRI, ACDC, and CHAOS [26]. Debiased-CAM studies image perturbation biases such as blur, color temperature, and day/night changes. It trains a multi-input, multi-task model with auxiliary explanation and bias-level prediction so that explanations remain closer to the unbiased scene [44]. These methods are important because they shift the goal from making a sharper heatmap to making the right heatmap.

Other hybrid methods focus on dense prediction. PCAA introduces Partial CAM for semantic segmentation and uses local and global class-level representations to model pixel-to-class relations [29]. RPIM builds foreground regions from superpixels and initial responses, then uses intra-region integration and inter-region spreading to improve consistency [46]. OLM refines low-level feature activation maps online to produce compact and threshold-robust localization maps [24]. These papers show that CAMs are not merely explanatory artifacts; in WSSS and WSOL they become training data.

CALM [14] takes a different route from post-hoc CAMs by embedding the attribution mechanism into the training process. Instead of generating a heatmap only after classification, it introduces a latent variable that represents the spatial location of the recognition cue and optimizes it with an expectation-maximization procedure. This makes the attribution map part of the model’s learning objective, which helps the classifier learn more localized and discriminative visual evidence. Its limitation is that it is not a purely post-hoc method and requires a modified training procedure.

## 5.3 High-resolution and fine-grained methods

High-resolution CAM methods address the spatial coarseness of final-layer activation maps. Augmented Grad-CAM [12] generates multiple low-resolution Grad-CAM maps from transformed copies of the same image and combines them into a super-resolution-style heatmap. Augmented Score-CAM [20] applies a similar augmentation idea to Score-CAM and reports better human preference, lower Average Drop, higher Average Increase, and better IoU than the original Score-CAM on ImageNet-based evaluations. F-CAM [18] attaches a trainable decoder to a classifier and uses foreground/background samples plus image priors to produce full-resolution CAMs. This improves boundaries, but it also makes the method less purely post-hoc because the decoder must be fine-tuned. LayerCAM [16] is one of the most influential high-resolution variants. It uses positive gradients at individual spatial locations and can generate reliable CAMs from different CNN layers. Shallow layers provide fine detail, while deeper layers provide semantic class evidence. By fusing these maps, LayerCAM improves WSSS and WSOL quality compared with single-layer CAMs. OLM [24] also uses low-level feature information, but it learns an online activation-map generator and evaluator. The goal is to produce a compact and threshold-robust map rather than a sparse discriminative patch. Poly-CAM [37] later combines earlier and later layers to produce higher-resolution maps that remain competitive on insertion-deletion faithfulness.

RPIM [46] introduces a region-based refinement mechanism. It first builds foreground regions from superpixels and initial CAM responses, then uses intra-region integration and inter-region spreading to make pixels inside a region and across related regions share more consistent activation. This is useful for WSSS because pixel-level pseudo-labels need coherent object regions rather than isolated peaks. GFR-CAM [52] uses Gram-Schmidt orthogonalization to generate a hierarchy of orthogonal explanation components. Unlike PCA-based approaches that focus on one dominant explanation, GFR-CAM can reveal secondary objects or semantic parts, reducing explanatory tunnel vision. Finer-CAM makes a different but very important observation: CAM often fails in fine-grained recognition because it explains what supports the target class, including features shared with similar classes. Finer-CAM instead compares the target class with similar reference classes and explains the logit difference [49]. Figure 6 shows that this comparison-based wrapper improves CUB localization scores for Grad-CAM, LayerCAM, and Score-CAM. The practical message is simple: for fine-grained explanations, the question should often be not 'why this class?' but 'why this class rather than the closest alternative?' Figure 6 shows that this comparison improves localization on CUB-200 and Cars when the same CAM backbone is evaluated with and without the Finer-CAM target. The remaining challenge is reference selection: the method is most effective when the reference class is truly visually competitive.

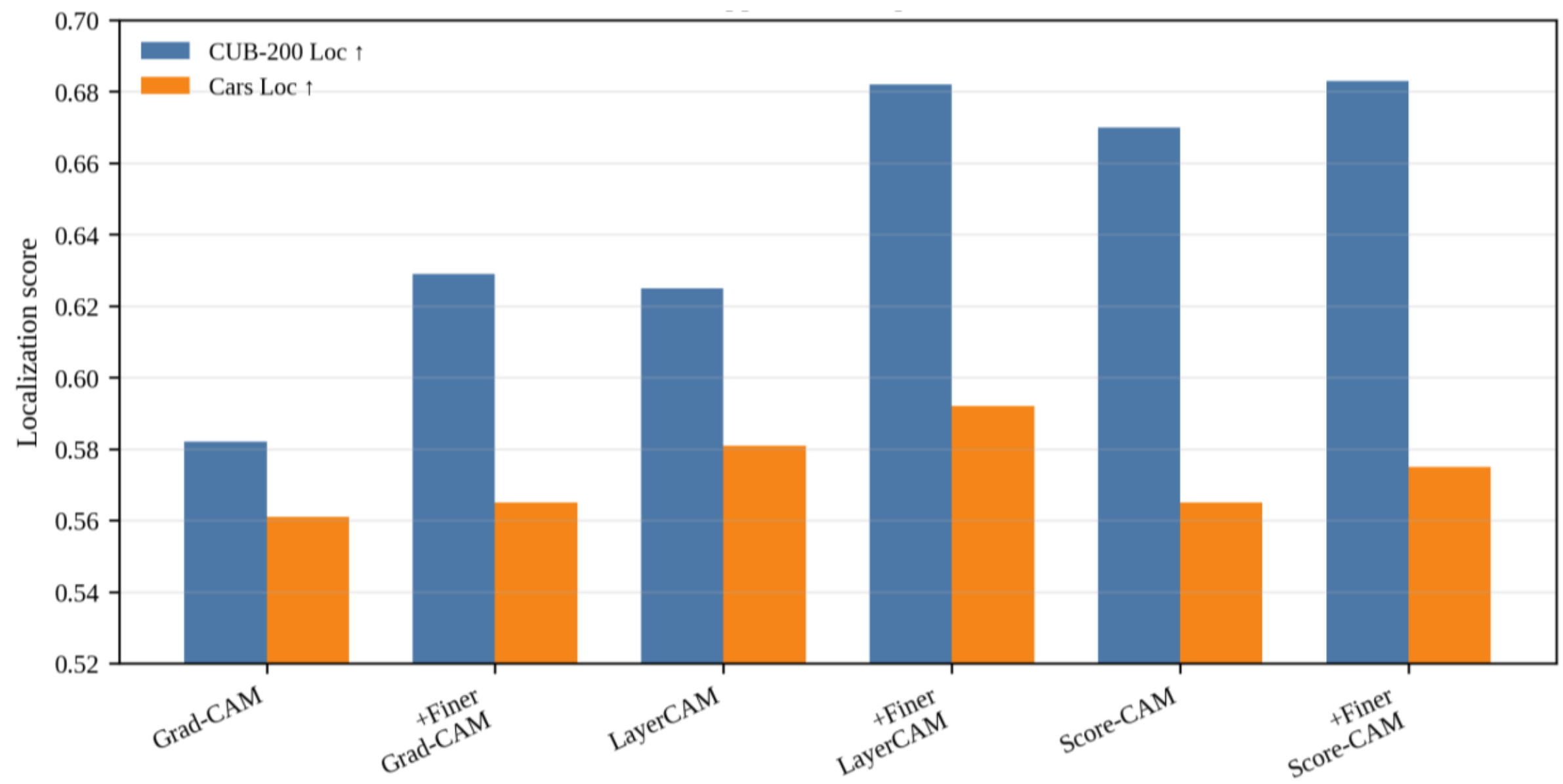

**Figure 6. Finer-CAM improves fine-grained localization by changing the explanation target from a class score to a target-versus-reference difference.**

## 5.4 Token-level and foundation-model-era CAMs

TS-CAM uses the self-attention mechanism of visual transformers to overcome the local-receptive-field limitation of CNN CAMs [31]. It couples token semantics with attention maps and reports a large Top-1 localization gain on CUB-200-2011. MCTformer extends this idea to WSSS by using multiple class tokens, so class-to-patch attention can generate class-specific localization maps [30]. CTI infuses class tokens within and across images and adds a background token to reduce false positives [41]. Prompt-CAM learns class-specific prompts for a pretrained ViT, making fine-grained traits visible through prompt-query attention [48]. TransCAM [34] extends CNN-based CAMs to hybrid CNN-Transformer models by aligning convolutional feature maps with transformer patch tokens. It integrates multi-level feature maps with class-token attention to produce high-resolution token-aware CAMs suitable for WSOL and fine-grained object localization. TransCAM demonstrates improved localization performance on CUB and Cars datasets while maintaining compatibility with pre-trained transformers.

Foundation-model methods broaden the target space. gScoreCAM explains CLIP by using gradients only to select the top channels and Score-CAM-style forward scores to weight them, reducing Score-CAM's CLIP runtime by about eight times while retaining strong object localization [47]. Figure 7 shows the reported COCO BoxAcc and runtime trade-off. S2C transfers SAM knowledge to the classifier during training through SAM-segment contrasting and CAM-based prompting [42]. The DINO semantic guider builds a class-aware affinity region map by propagating CAM seeds over DINO self-attention affinity graphs [43]. DiffCAM explains a target example by comparing its feature distribution with reference examples rather than relying on decision-boundary gradients [50]. MetaCAM combines multiple CAM methods through top-k consensus and adaptive thresholding, showing that ensemble agreement can outperform individual maps [54].

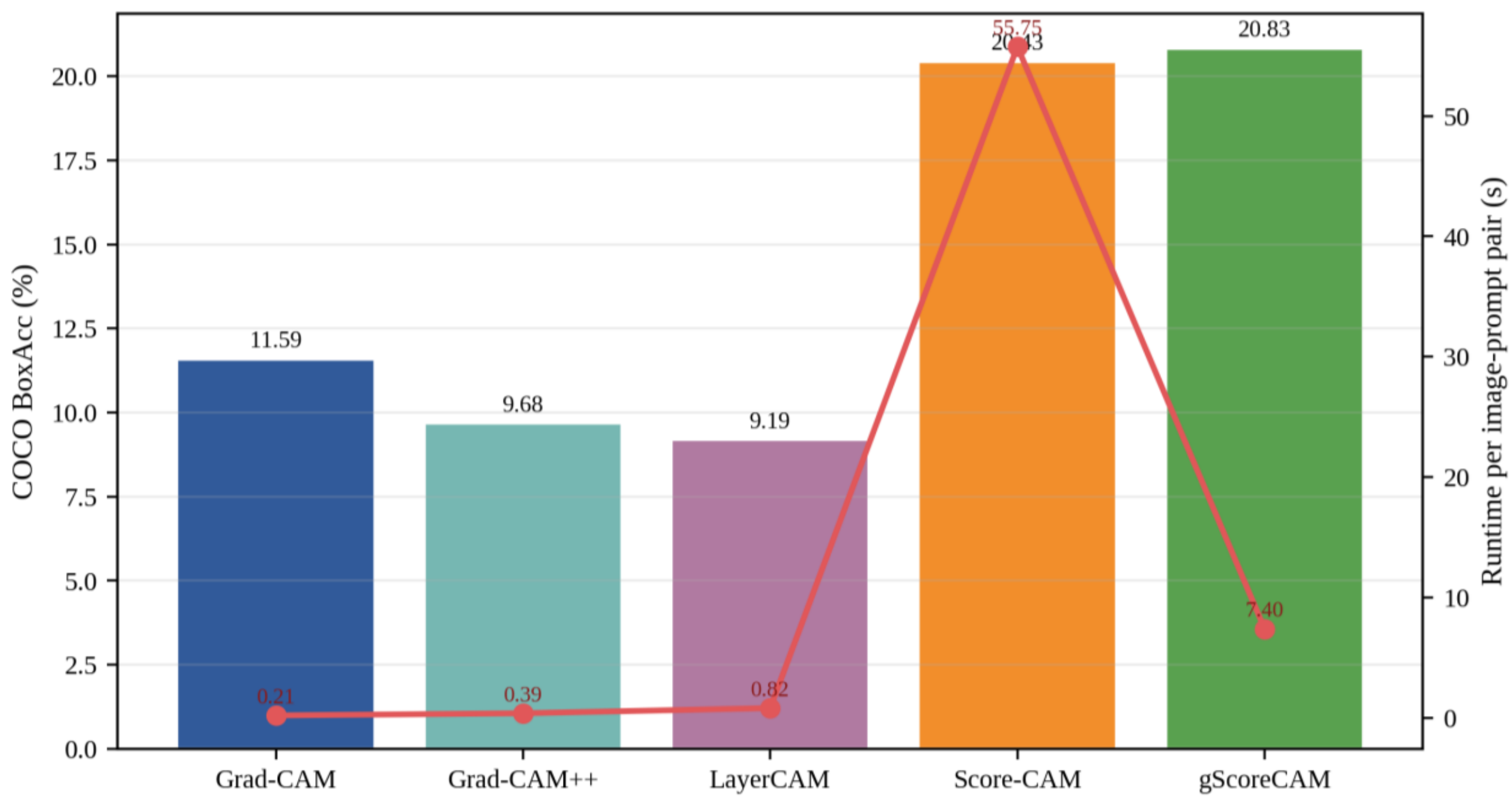

**Figure 7. CLIP localization and runtime trade-off reported in gScoreCAM. Score-CAM and gScoreCAM localize well, but gScoreCAM greatly reduces runtime.**

**Table 6. Summary of recent, hybrid, gradient-free, high-resolution, and foundation-model-era CAM-style methods.**

| Method | Core mechanism | Main novelty | Remaining gap | Datasets / reported result |
|---|---|---|---|---|
| Score-CAM [8] | Forward score weighting with activation masks. | Gradient-free explanation and sanity-check support. | High cost from many forward passes. | ImageNet and localization tasks; cleaner maps than gradient CAMs. |
| Ablation-CAM [9] | Remove feature maps and measure score drop. | Direct marginal contribution estimate. | Requires many ablations. | Localization and trust evaluation. |
| Ablation-CAM++ [19] | Grouped recursive ablations. | Reduces Ablation-CAM cost. | Grouping can smooth fine details. | ICIP 2022 efficiency-focused evaluation. |
| Eigen-CAM [21] | Principal component of feature maps. | No gradients and no correct class requirement. | Less class-specific. | WSOL, segmentation, object proposal gains. |
| Cluster-CAM [38] | Cluster feature maps before forward scoring. | Efficient gradient-free CAM. | Cluster quality affects maps. | ImageNet evaluations; faster than dense score methods. |
| CAPE [39] | Probabilistic ensemble reformulation. | Comparable regional contributions across classes. | Requires probabilistic modeling assumptions. | CUB, ImageNet, and CMML cytology. |
| ReciproCAM [40] | Spatial perturbation of feature maps. | Fast gradient-free explanations. | Target-layer choice controls resolution/cost. | ILSVRC2012; strong ADCC and 148x faster than Score-CAM. |
| ScoreCAM++ [51] | Gated score weighting with tanh normalization. | Improves Score-CAM faithfulness. | Still forward-score based. | ImageNet and Cat/Dog; lower Average Drop and higher confidence. |
| SEAM [11] | Equivariant CAM regularization and pixel correlation. | More consistent WSSS seeds. | Training-specific pipeline. | PASCAL VOC pseudo labels improved to 55.41% mIoU before CRF. |
| AdvCAM [23] | Anti-adversarial input manipulation. | Expands attributions beyond most discriminative parts. | Requires iterative manipulation and regularization. | PASCAL VOC WSSS and semi-supervised segmentation. |

| Method | Core mechanism | Main novelty | Remaining gap | Datasets / reported result |
|---|---|---|---|---|
| C2AM [25] | Contrastive class-agnostic activation maps. | Foreground-background disentanglement without labels. | Class-agnostic maps need refinement for class labels. | CUB, ImageNet, PASCAL VOC; improves initial CAMs. |
| C-CAM [26] | Causal category and anatomy chains. | Addresses medical boundary ambiguity and co-occurrence. | Domain-specific causal assumptions. | ProMRI, ACDC, CHAOS; high pseudo-mask DSC. |
| Debiased-CAM [44] | Multi-task debiasing for image perturbations. | Improves explanation faithfulness under bias. | Requires bias labels or simulated bias levels. | Blur, color temperature, day/night; user studies. |
| CI-CAM [45] | Causal intervention on object-context confounding. | Reduces biased object-context entanglement. | Backbone and intervention design matter. | CUB Top-1 Loc 58.39%; ImageNet comparable. |
| F-CAM [18] | Trainable guided upscaling decoder. | Full-resolution CAMs. | Fine-tuning decoder adds complexity. | CUB and OpenImages WSOL. |
| LayerCAM [16] | Layer-wise positive gradient-weighted activation. | Combines coarse semantic and shallow spatial details. | Fusion choices matter. | PASCAL VOC val/test mIoU 60.8/61.4 with VGG16. |
| RPIM [46] | Superpixel region integration and spreading. | Improves pseudo-label consistency. | Depends on superpixel quality. | PASCAL VOC WSSS; plug-and-play improvements. |
| Finer-CAM [49] | Target-vs-reference logit difference. | Fine-grained discriminative trait localization. | Reference-class selection matters. | Birds, CUB, Cars, Aircraft, FishVista. |
| gScoreCAM [47] | Gradient-guided channel selection for CLIP Score-CAM. | Much faster CLIP localization than Score-CAM. | Still needs selected forward passes. | ImageNet-v2, COCO, PartImageNet. |
| DiffCAM [50] | Feature-difference saliency from data distributions. | Works without decision-boundary gradients. | Reference distribution choice matters. | RSNA/SIIM medical AUPRC and SSL/ViT case studies. |
| MetaCAM [54] | Consensus ensemble of CAM methods. | Improves robustness and ROAD score. | Depends on component method set. | Large-scale ensemble experiments. |
| Poly-CAM [37] | Multi-layer fusion of early and late convolutional features | Produces high-resolution CAMs while preserving competitive insertion/deletion faithfulness. | Fusion design and layer selection remain architecture-dependent | CUB/OpenImages or ILSVRC evaluations. |
| CALM [14] | Integrates attribution into the training process via a latent spatial variable | Embeds attribution map learning in classifier training using expectation-maximization; improves localization and discriminative evidence coverage | Requires modified training; not purely post-hoc; may not generalize across architectures | Reported on PASCAL VOC / ImageNet for WSOL; improves localization mIoU |

Table 6 is intentionally compact, but it shows the diversity of the recent literature. The common pattern is that each method solves one bottleneck while opening another. Score-CAM removes gradients but increases cost. LayerCAM increases resolution but requires layer fusion. Finer-CAM improves fine-grained discrimination but needs a reference class. gScoreCAM adapts Score-CAM to CLIP but relies on channel selection. DiffCAM avoids decision-boundary gradients but depends on reference data.

## 6. Cross-family Comparison: Post-hoc Gradient-based versus Gradient-free/Hybrid CAMs

This section compares only post-hoc gradient-based methods and gradient-free or hybrid CAM methods when the source papers report compatible values. It deliberately excludes the CNN-specific training and architecture-aware methods reviewed in Section 7. The reason is methodological: Hide-and-Seek, ACoL, SPG, ADL, CREAM, and related methods change the training pipeline or localization architecture, so their WSOL numbers should not be mixed with post-hoc explanation scores. Section 6 therefore focuses on three questions that can be compared more fairly: runtime, faithfulness, and localization under shared post-hoc protocols.

Figure 8 compares runtime between some of the post-hoc gradient-based methods and gradient-free or hybrid CAM methods, and includes only methods from the strict corpus that are reported in the papers with same runtime status reported in [47]. Grad-CAM and Grad-CAM++ are fast because they use one backward pass. Score-CAM is much slower because it requires thousands of forward passes in the CLIP setting. gScoreCAM keeps the Score-CAM scoring idea but selects only a subset of channels, producing a much smaller runtime while preserving localization quality

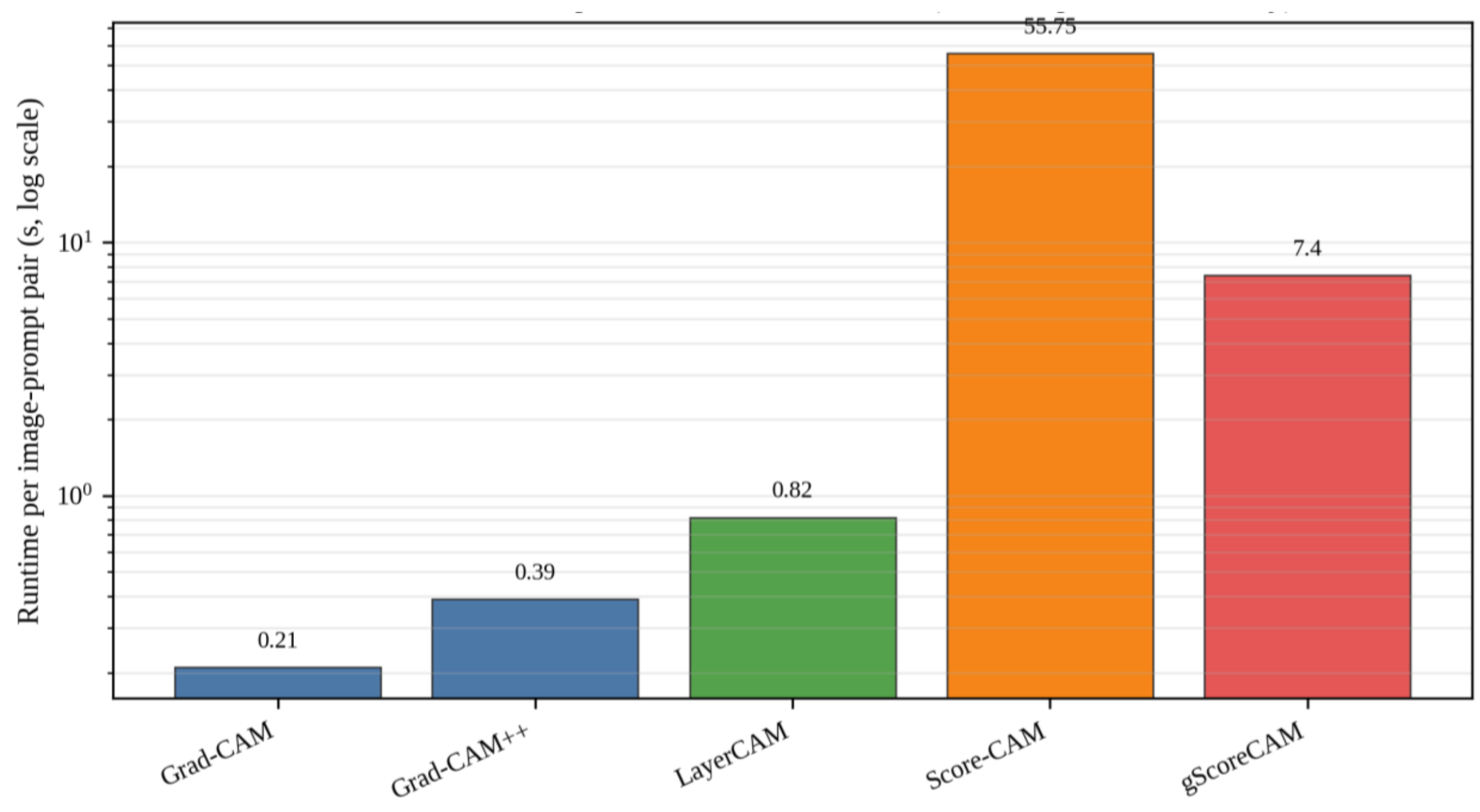

**Figure 8. Runtime comparison between strict-corpus post-hoc methods.**

Figure 9 provides a faithfulness trend under a more controlled setting that the Poly-CAM paper reports insertion and deletion for several CAM-style methods on ILSVRC2012 / VGG16. The graph shows two lessons. First, progress is not monotonic; a newer method is not automatically better on both insertion and deletion. Second, high-resolution or carefully fused methods such as LayerCAM and Poly-CAM can improve deletion while maintaining strong insertion, suggesting that spatial detail and faithfulness do not always conflict.

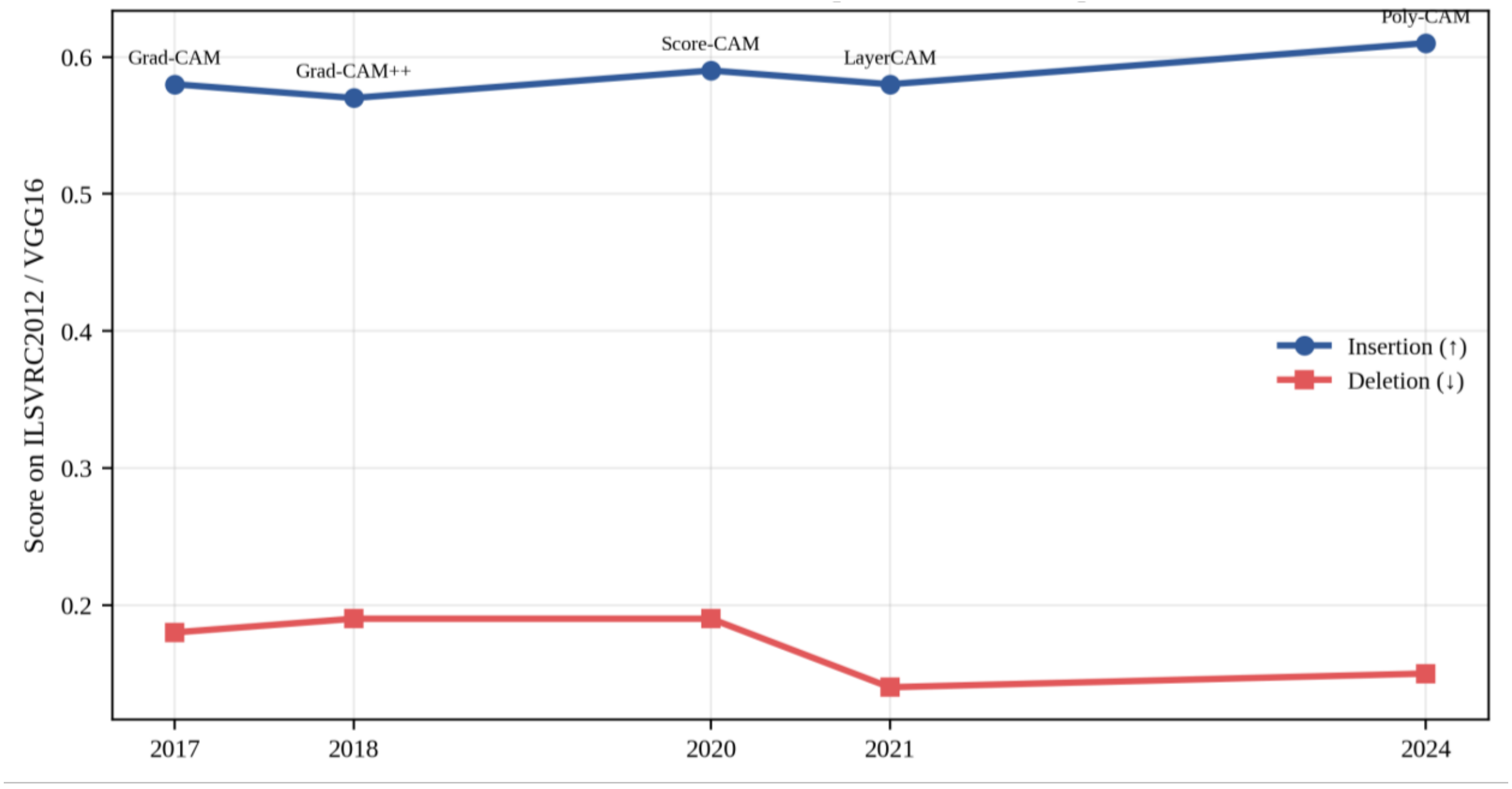

**Figure 9. Faithfulness trend under a single reported ILSVRC2012 / VGG16 protocol. Insertion should increase and deletion should decrease.**

Figure 10 uses the LIFT-CAM protocol to compare localization through the proportion of explanation energy inside ImageNet bounding boxes. LIFT-CAM is strongest among the shown methods, while Ablation-CAM and Score-CAM also exceed Grad-CAM and Grad-CAM++. This does not mean that gradient-free methods are always better; it means that direct marginal scoring and additive attribution can place evidence more compactly in the object region under this specific evaluation.

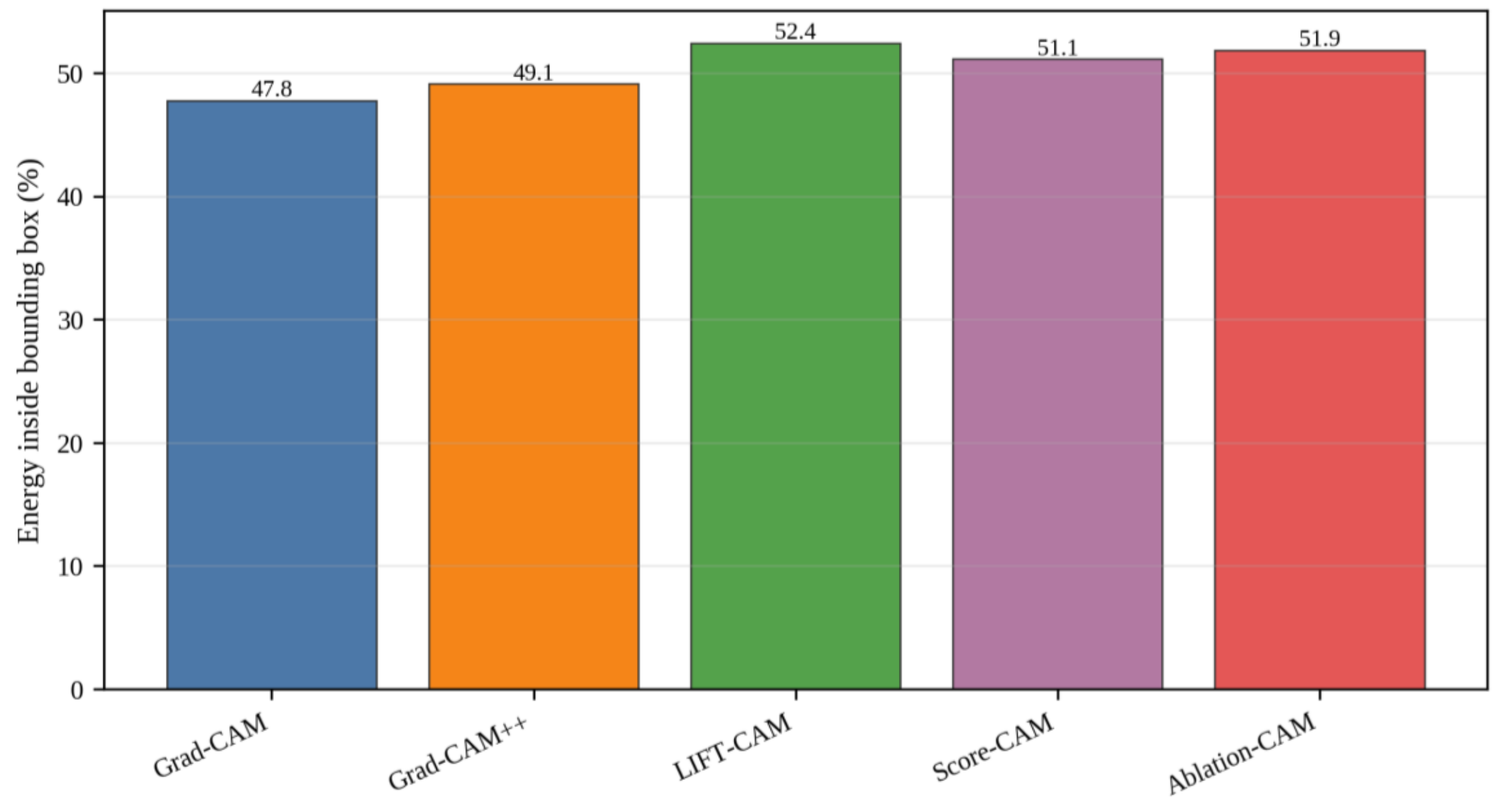

**Figure 10. Localization comparison between gradient-based and gradient-free post-hoc CAMs under the same ImageNet energy-localization protocol.**

The cross-family comparison leads to three practical conclusions. First, Grad-CAM-style methods remain the fastest and most convenient first-line diagnostic tools. Second, gradient-free and additive-attribution methods often improve faithfulness or localization, but their computational cost or design complexity can be higher. Third, high-resolution and foundation-model methods should be evaluated with special care because their maps may reflect external priors, prompts, or learned reference distributions rather than only the original classifier.

## 7. Model-based and Architecture-aware CAM Methods

Model-based and architecture-aware methods do not treat CAM as a purely post-hoc heatmap. They modify training, pooling, erasing, guidance, attention, or class-specific prototypes so that the model itself becomes more localizable. These methods are reviewed separately from Section 6 because their numbers often reflect a different learning pipeline rather than a direct post-hoc explanation algorithm.

### 7.1 CNN-specific localization mechanisms

The original CAM method [1] is architecture-aware because it depends on global average pooling followed by a linear classifier. This design preserves the spatial meaning of the last convolutional maps and makes it possible to project class weights back to image regions. Hide-and-Seek [4] modifies the training image by randomly hiding patches. The model is forced to use less discriminative object parts when the most discriminative part is hidden. ACoL [5] uses two classifiers and adversarial erasing inside the feature map: one branch finds discriminative regions, and the other branch learns complementary regions after those regions are suppressed. SPG [6] turns high-confidence attention regions into self-produced foreground/background guidance masks, gradually adding spatial supervision during classification training. ADL [7] uses an attention-based dropout layer that sometimes erases the most discriminative region and sometimes highlights informative regions, balancing localization completeness and recognition accuracy.

Rethinking CAM [10] analyzes the standard WSOL pipeline and identifies three underappreciated causes of poor localization: global average pooling can overemphasize channels with small activation areas, negative weights can suppress true object regions, and max-based thresholding can be unstable. It proposes thresholded average pooling, negative weight clamping, and percentile thresholding. CREAM [27] addresses incomplete localization by modeling foreground and background activation distributions. It learns class-specific context embeddings and then performs class re-activation through a Gaussian-mixture-style estimation. LPCAM [35] attacks the discriminative-feature problem by extracting local prototypes from non-discriminative object parts. Instead of using only classifier weights, it clusters unpooled local features for an object class and matches new images to those prototypes, covering object parts such as head, body, and legs.

## 7.2 Transformer and token-aware mechanisms

Although token-level methods were reviewed in Section 5, they are also architecture-aware because they exploit the internal structure of transformers. TS-CAM [31] uses long-range self-attention to reduce partial activation. MCTformer [30] introduces multiple class tokens so that different class-to-patch attentions can produce class-specific maps. CTI [41] infuses class tokens from same-image and cross-image views to improve global-local consistency and class specificity. Prompt-CAM [48] learns class-specific prompts for a pretrained ViT, making the attention heads reveal fine-grained traits rather than only the whole object. These methods show that the transformer era changes the unit of explanation from convolutional channels to token interactions.

**Table 7. Model-based and architecture-aware CAM methods.**

| Method | Architecture-aware mechanism | Contribution | Remaining limitation |
|---|---|---|---|
| CAM [1] | GAP + linear classifier | Projects classifier weights onto convolutional maps. | Requires a compatible architecture. |
| Hide-and-Seek [4] | Randomly hidden input patches | Forces the model to seek alternative object parts. | Training-time augmentation; not post-hoc. |
| ACoL [5] | Parallel classifiers with adversarial erasing | Learns complementary regions in feature space. | Erasing design affects coverage. |
| SPG [6] | Self-produced foreground/background guidance | Adds spatial correlation supervision from high-confidence regions. | Stagewise training and threshold choices. |
| ADL [7] | Attention-based dropout/highlighting | Balances erasing and attention strengthening. | Classification-localization trade-off. |
| Rethinking CAM [10] | Pooling and thresholding corrections | Fixes GAP bias, negative weights, and unstable thresholds. | Still relies on CAM pipeline. |
| CREAM [27] | Class re-activation with foreground/background embeddings | Separates less-discriminative foreground from background. | Requires context-embedding learning. |
| LPCAM [35] | Local prototypes for non-discriminative features | Covers local semantics discarded by discriminative classifiers. | Prototype selection and class coverage. |
| TS-CAM [31] | Token semantic coupled attention | Uses ViT long-range dependency for WSOL. | Class-token semantics must be learned. |
| MCTformer [30] | Multiple class tokens | Class-specific class-to-patch attention for WSSS. | Token-to-mask refinement needed. |
| CTI [41] | Intra-/cross-image class token infusion | Improves token consistency and class specificity. | Additional token-infusion training. |
| Prompt-CAM [48] | Class-specific prompts for ViTs | Fine-grained trait localization from prompt attention. | Prompt design and head importance matter. |

Table 7 shows that architecture-aware methods usually improve localization by changing the model's learning process or exploiting architectural tokens. The trade-off is reduced generality: a method designed for GAP CNNs, dual-branch classifiers, or ViT class tokens may not transfer directly to another architecture.

# 8. Comparative Discussion

The first lesson from the corpus is that CAM-style methods now answer several different questions. Grad-CAM asks which feature maps locally support a score [2]. Score-CAM asks which activation masks preserve the score under forward evaluation [8]. Finer-CAM asks which regions distinguish one class from its nearest alternatives [49]. FAM asks what part of an image contributes to a representation rather than to a class label [28]. SAM- and DINO-assisted methods ask how external pretrained structure can improve weakly supervised CAM seeds [42], [43]. These are related but not identical explanation problems.

The second lesson is that faithfulness and localization should not be conflated. A map can overlap an object and still be unfaithful to the model if the model actually relied on background context. Conversely, a faithful map may reveal an unwanted shortcut. This is why causal and debiasing methods are important. CI-CAM, C-CAM, and Debiased-CAM show that the quality of an explanation depends not only on the heatmap formula but also on the causal structure and bias conditions of the data [26], [44], [45].

The third lesson is that resolution must be controlled semantically. LayerCAM, F-CAM, Poly-CAM, RPIM, and GFR-CAM improve spatial detail [16], [18], [37], [46], [52]. However, a high-resolution map is not automatically a better explanation. If the target score is broad or confounded, higher resolution may simply reveal the wrong evidence in greater detail. Finer-CAM is valuable because it changes the target to a contrastive question, making high-resolution detail more meaningful in fine-grained recognition [49].

The fourth lesson is that foundation models complicate explanation ownership. If SAM improves CAMs, the final map partly reflects SAM's segmentation prior [42]. If DINO affinities spread CAM seeds, the final pseudo-label partly reflects DINO's self-supervised representation [43]. If CLIP is explained with a prompt, the heatmap depends on the wording and embedding of that prompt [47]. Future CAM papers should therefore report not only the heatmap method but also the source of all external priors and prompts.

## 9. Open Challenges and Future Directions

A first challenge is standardized evaluation. Many CAM papers use different target layers, input sizes, thresholds, perturbation baselines, and post-processing. Future papers should report a minimal evaluation card: model and layer, target definition, normalization, thresholding, number of forward and backward passes, whether CRF/SAM/saliency priors were used, and which metrics were computed. This would make cross-paper comparison more reliable.

A second challenge is causal faithfulness. Confounding is common in natural images, and medical images often contain anatomical co-occurrence. CI-CAM and C-CAM provide early examples of causal reasoning in CAM-style methods [26], [45]. Future work should combine interventional datasets, counterfactual image editing, and causal metrics to test whether highlighted evidence is necessary for the model decision or merely correlated with it.

A third challenge is efficient gradient-free explanation. Score-CAM and Ablation-CAM are attractive because they avoid unstable gradients, but their cost limits deployment [8], [9]. ReciproCAM, Cluster-CAM, AD-CAM, and Ablation-CAM++ show that structured perturbation, grouping, and feature-level masking can reduce this cost [19], [36], [38], [40]. A promising direction is adaptive explanation: use fast gradients when they are reliable, and switch to perturbation only when uncertainty or inconsistency is high.

A fourth challenge is prompt and reference sensitivity in foundation models. Finer-CAM depends on the reference class, gScoreCAM depends on prompt-conditioned CLIP scores, and DiffCAM depends on reference data distributions [47], [49], [50]. Future evaluations should include prompt sensitivity, reference-class sensitivity, and reference-set sensitivity as standard robustness tests.

A fifth challenge is human validation. Human studies in Grad-CAM and Debiased-CAM show that explanations can help users calibrate trust and improve task performance [2], [44]. Yet human preference should not be the only criterion because visually pleasing maps can still be unfaithful. Future work should combine human-centered evaluation with deletion/insertion, sanity checks, causal interventions, and task outcomes.

## 10. Conclusion

CAM-style explanation has evolved from a simple CNN localization mechanism into a broad methodological family for explainable computer vision. The original CAM formulation showed that a classifier can reveal discriminative regions through global average pooling [1]. Grad-CAM made this idea broadly applicable [2]. Later work improved derivatives, avoided gradients, refined resolution, used causal and debiasing constraints, introduced token-level explanations, and incorporated foundation-model priors.

The main conclusion of this review is that CAMs should not be treated as interchangeable heatmaps. Each method answers a specific explanatory question with a specific technical mechanism. A fast-debugging tool, a faithful post-hoc explanation, a dense pseudo-label generator, a fine-grained trait localizer, and a CLIP prompt explanation are related but different objects. A rigorous CAM study should therefore report the target being explained, the evidence source, the attribution mechanism, the evaluation protocol, the computational cost, and the limitations of the explanation. This method-centered view is essential for building a more reliable and useful next generation of visual explanations.